\documentclass[10pt,letterpaper]{article}

\usepackage{cogsci}
\usepackage{pslatex}
\usepackage{apacite}

\usepackage{graphicx}

\title{Modeling Human Inference of Others' Intentions in Complex Situations \\ with Plan Predictability Bias}
 
\author{{\large \bf Ryo Nakahashi (ryon@nii.ac.jp)} \\
  The Graduate University for Advanced Studies(Sokendai) \\
  2-1-2 Hitotsubashi, Chiyoda, Tokyo, Japan
  \AND {\large \bf Seiji Yamada (seiji@nii.ac.jp)} \\
  National Institute of Informatics \\
  The Graduate University for Advanced Studies (Sokendai) \\
  2-1-2 Hitotsubashi, Chiyoda, Tokyo, Japan }

\begin{document}

\maketitle

\begin{abstract}
A recent approach based on Bayesian inverse planning for the ``theory of mind'' has shown good performance in modeling human cognition. 
However, perfect inverse planning differs from human cognition during one kind of complex tasks due to human bounded rationality. One example is an environment in which there are many available plans for achieving a specific goal. 
We propose a ``plan predictability oriented model'' as a model of inferring other peoples' goals in complex environments. This model adds the bias that people prefer “predictable” plans. This bias is calculated with simple plan prediction. We tested this model with a behavioral experiment in which humans observed the partial path of goal-directed actions. Our model had a higher correlation with human inference. We also confirmed the robustness of our model with complex tasks and determined that it can be improved by taking account of individual differences in ``bounded rationality''.

\end{abstract}

\section{Introduction}

People have a cognitive mechanism called the ``theory of mind'' that can estimate peopleâ€™s purposes and plans through observation from infancy \cite{Woodward2001}. Trials for using this theory of mind as a computational model are of great interest in the cognitive science arena, and there has been a lot of research about this \cite{Goldman2012} \cite{Paul2006}.

A recent approach based on Bayesian inverse planning has been recognized as a method for modeling the theory of mind  \cite{Baker2009}. In this approach, people are modeled as agents who are rational \cite{Dennett1987}. When a goal is given, rational agents behave rationally to achieve it. The posterior probability of a particular goal given observed behavior is calculated by the product of the prior probability of the goal and the likely behavior of the rational agent given the goal.

This approach can be applied to all problems in which rational agents can be designed. There are many examples of functions used to estimate other peoples' intentions on the basis of this approach, such as other peoples' intermediate goals \cite{Nakahashi2015}, preferences toward navigation \cite{Jara-ettinger2015}, 
and so on.

However, this approach may differ from human perception in some complex situations.


\begin{figure}[t]
  \begin{center}
   \includegraphics[width=\linewidth]{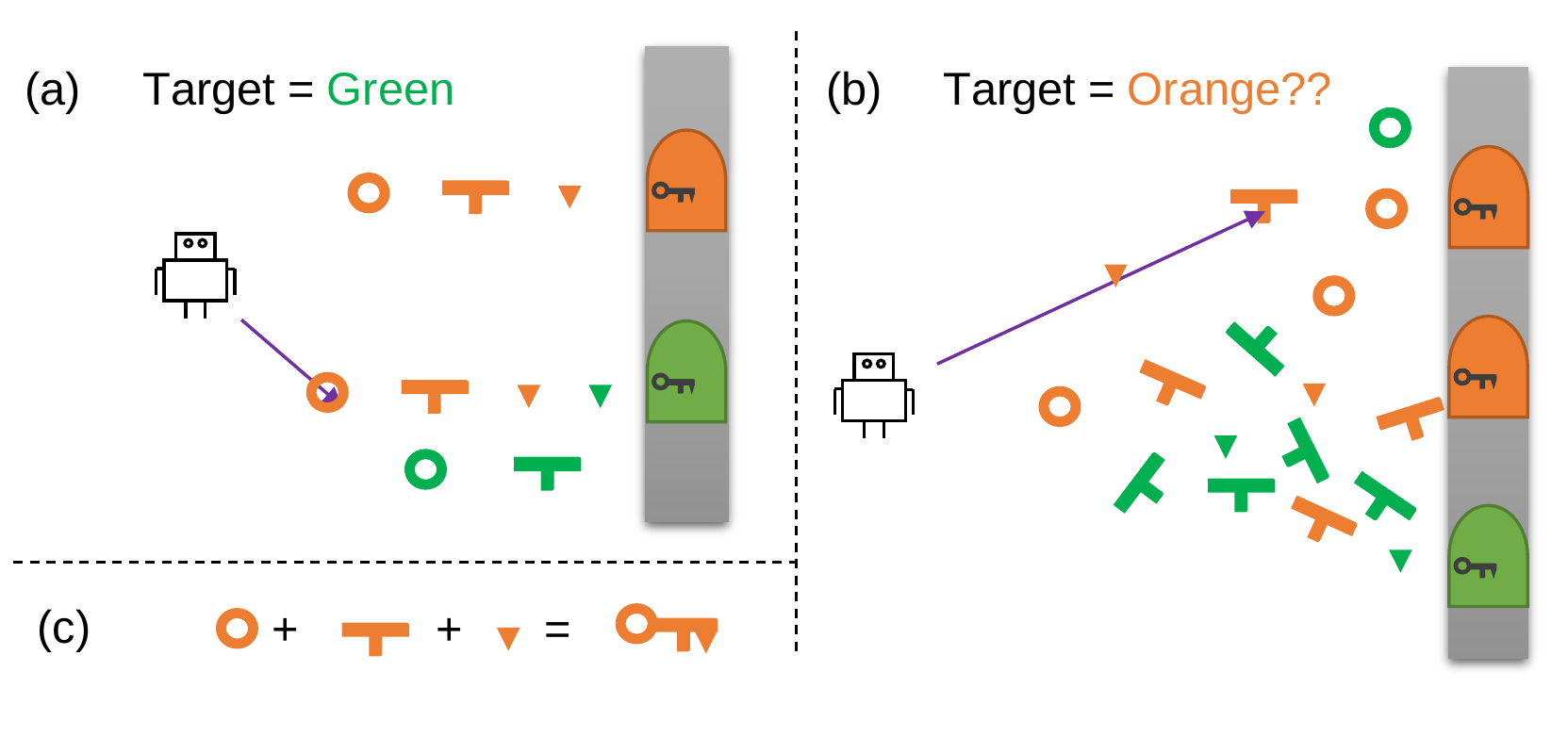}
  \end{center}
 \caption{Complex environment in which Bayesian inverse planning is mistaken for model human inference. Bayesian inverse planning infers green door as robots goal in (a), (b). Humans may infer orange door in (b). (c) is rule for assembling keys.}
 \label{fig:example}
\end{figure}

Figure \ref{fig:example} is one example of such a problem. In Figure \ref{fig:example} (a), the robot wants to open and go through one of the doors. It wants to go through either the orange door or the green door, and the doors require a key of the same color to be opened. To open a door, the robot assembles the required key. To do so, it is necessary to collect three types of parts as shown in Figure \ref{fig:example} (c). In \ref{fig:example} (a), the robot moved as indicated by the purple line. Which door is the robot about to open, the orange or the green one? In this case, many people will answer with the green one. This is because, if the robot wanted to open the orange door, it would collect the three parts shown on the upper half and open the orange door. How about the example in \ref{fig:example} (b)? Perfectly rational thought leads to the conclusion that the correct answer is green, as in \ref{fig:example} (a). The reason is that there is a shorter path to collect the orange key parts than the current robot path. For this reason, the inference formed on the basis of the Bayesian inverse planning model will be that the robot wants to go through the green door in this example. However, we often infer that the robot wants to go through the orange door when looking at Figure \ref{fig:example} (b).

This difference is due to the fact that observers cannot completely recognize the rationality of the actors when a problem is complicated. In the above example, there are many possible plans to achieve one goal, but it is difficult to evaluate all of them. On the basis of this reason, human inference is affected by the bias towards ``easy to predict'' plans. This means that humans only consider a few plans that can be easily predicted. This is an example of bounded rationality  \cite{Simon1957}.

In this paper, we propose a novel model called the ``plan predictability oriented model'' for modeling human inference with plan predictability bias. We aim to build a more adequate model of human inference in problem settings in which there are many plans available to achieve one goal, as shown in the example. This model is based on Bayesian inverse planning, but we allowed for bias in plan predictability in the inference phase. We perform calculation with the likelihood of a future plan from observed behavior like plan prediction 
\cite{Charniak1993}, and we use the likelihood as our plan predictability bias. We use simple soft-max likelihood base plan prediction.

To show the advantages of our proposed model, we designed a scenario called ``item creating'' and carried out subject experiments with it. In this scenario, there is an agent that collects parts to create a specific item. Participants observe part of the behavior of the agent and are then asked to estimate the item that the agent wants to create on the basis of the observation. We compared the correlation between this participantâ€™s data and both the inferences formed on the basis of the full inverse planning model and our model. The result was that our model had a better correlation with human inference than did the full inverse planning model. Additionally, we confirmed that a higher task complexity made the accuracy of the full inverse planning model worse, but our model was not affected. We also show that there are individual differences regarding the bias and that we can improve our model by considering individual differences.

We wrote a problem setting, detailed the full inverse planning model and plan predictability oriented model, wrote the details and results of evaluations with the ``item creating'' scenario, discussed and summarized our approach.

\section{Computational Model}

\subsection{Preliminary}
\subsection{Notation and Problem Setting}

We denote the set of people's goals as $\mathcal{G}$ and the set of actions as $\mathcal{A}$.
Action sequences are represented as $\mathbf{a}\in\mathcal{A}^+$, 
and the set of all plans to achieve goal $g\in\mathcal{G}$ are represented as $\mathcal{P}_g$.
Since plans are a kind of sequence, $\mathcal{P}_g\subset\mathcal{A}^+$.

\subsection{Model of Human Rationality}
We use Boltzmann noisy rationality as the rationality of action for our model. In the definition for Boltzmann noisy rationality, $P(a|g)$ is the probability that an agent executes $a$ for $g$ in accordance with the Boltzmann distribution of the ``value'' of $a$ for achieving $g$. Equation \ref{eq:rationality} is the definition of the probability. $\beta$ is the temperature parameter of the Boltzmann distribution to define rationality.
ă
\begin{eqnarray}
\label{eq:rationality}
P(g|a) = \frac{\exp{(\beta Q_g(a))}}{\sum_{g'\in \mathcal{G}} \exp{(\beta Q_{g'}(a))}}
\end{eqnarray}
Here, $Q_g(a)$ corresponds to the value function in terms of MDP planning or reinforcement learning 
which means expected rewards after executing $a$ under $g$.

\subsection{Modeling and Calculation}

\subsection{Problem Objective}

The objective of our modeling is the posterior probability that the goal of others is $g$ when observing others action sequence ${\bf a}$.
\begin{eqnarray}
\label{eq:object}
P(g|\mathbf{a})
\end{eqnarray}

\subsubsection{Calculating Full Inverse Planning Model}
In the inverse planning approach, we reverse the variable dependency in Eq. \ref{eq:object} by using a Bayesian theorem.
\begin{eqnarray}
\label{eq:full_1st}
P(g|\mathbf{a}) \propto P(\mathbf{a}|g)P(g)
\end{eqnarray}

We assume no prior knowledge about another persons goal. In other words, we assume $P(g)$ as a uniform distribution. Therefore, we can ignore $P(g)$ from Eq. \ref{eq:full_1st}. To calculate this, we assume that a human creates a plan in advance and executes their actions according to the plan. Thus, we can factorize $P(\mathbf{a}|g)$ into the probability of a plan to be considered as one achieving $g$ and the probability of executing $\mathbf{a}$ under the plan. The following equation is obtained when summarizing this factored probability for all available plans.

\begin{eqnarray}
\label{eq:full_bayes}
P(\mathbf{a}|g) \propto \sum_{p \in \mathcal{P}_g}{P(\mathbf{a}|p)P(p|g)}
\end{eqnarray}

$P(\mathbf{a}|p)$ and $P(p | g)$ are calculated using Boltzmann noisy rationality.

\begin{eqnarray}
\label{eq:full_Q}
P(\mathbf{a}|p) = \frac{\exp{(\beta Q_{p, g}(\mathbf{a}))}}{\sum_{p' \in \mathcal{P}_g} \exp{(\beta Q_{p', g}(\mathbf{a}))}} \nonumber\\ 
P(p|g) = \frac{\exp{(\beta Q_g(p))}}{\sum_{g'\in \mathcal{G}} \exp{(\beta Q_{g'}(p))}}
\end{eqnarray}
Note, that $p$ is a plan for achieving a specific goal. In other words, if a plan decided, the corresponding goal also comes uniquely. Thus we can treat $ Q_{p, g}(\mathbf{a})$ as $Q_p(\mathbf{a})$

\subsubsection{Calculating Plan Predictability Oriented Model}

In the plan predictability oriented model, Eq. \ref{eq:full_1st} is the same; however, the way of calculating of $P(\mathbf{a}|g)$ is made different in order to integrate the bias that people prefer predicable plans. The predictability of a plan is the probability that a plan is executing from observing action sequences, thus that is $p(p|\mathbf{a})$. We use this  instead of $P(\mathbf{a}|p)$; thus, we obtain the following equation.

\begin{eqnarray}
\label{eq:ppom}
P(\mathbf{a}|g) \propto \sum_{p \in \mathcal{P}_g}{P(p|\mathbf{a})P(p|g)}
\end{eqnarray}

Here, we use simple Boltzmann noisy rationality for $P(p|\mathbf{a})$ as follows.

\begin{eqnarray}
\label{eq:ppom_Q}
P(p|\mathbf{a})  = \frac{\exp{(\beta Q_{p}(\mathbf{a}))}}{\sum_{\mathbf{a} \in \mathcal{A}^+} \exp{(\beta Q_{p}(\mathbf{a}))}}
\end{eqnarray}

\subsubsection{Comparison of Full Inverse Planning Model and Plan Predictability Oriented Model}

\begin{figure}[t]
  \begin{center}
   \includegraphics[width=0.7\linewidth]{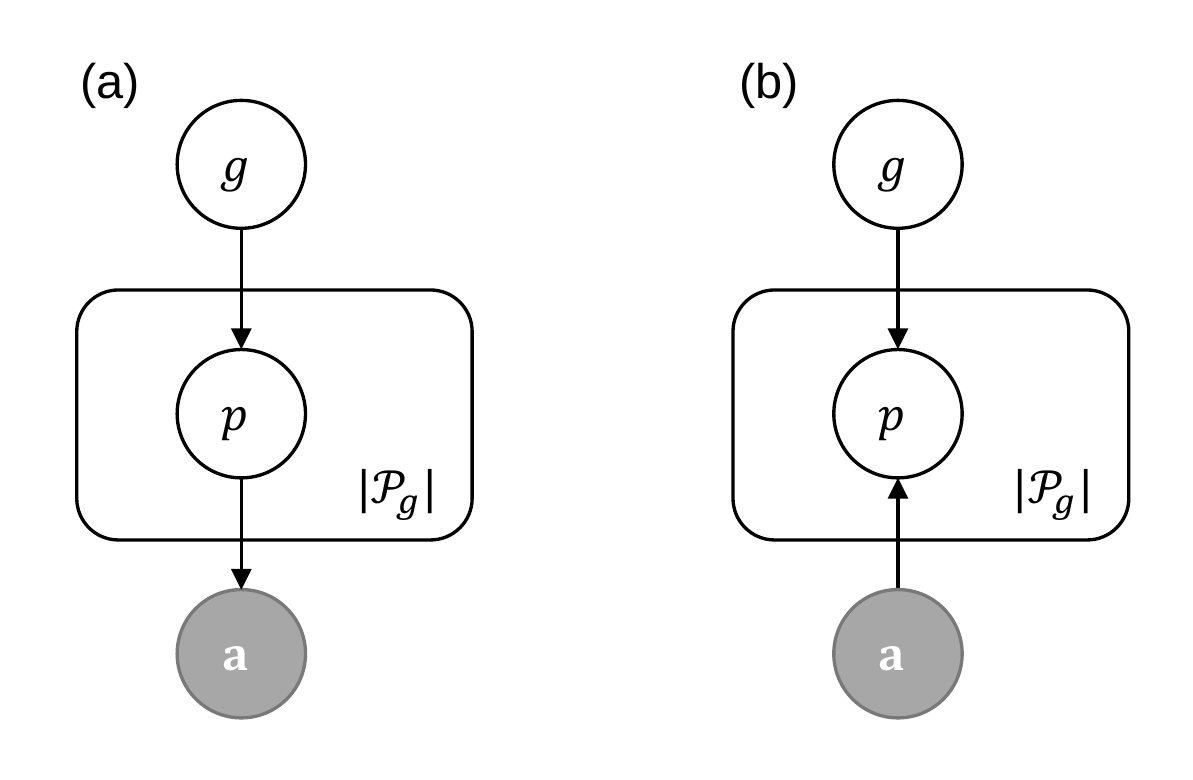}
  \end{center}
 \caption{Graphical model of $P(\mathbf{a}|g)$ for (a) full inverse planning model and (b) plan predictability oriented model.}
 \label{fig:graphical}
\end{figure}

Figure \ref{fig:graphical} shows the difference in graphical models for $P(g|\mathbf{a})$. As shown Figure \ref{fig:graphical} (a), in the full inverse planning model, humans assume that others decide their plan in advance and then act according to it. This is the natural process of human planning. In comparison, as Figure \ref{fig:graphical} (b), in the plan predictability oriented model, action sequence affect plans. This means that humans may change their plans depending on their actions, and this is unintuitive. 
We assume that humans cannot calculate the goals of others on the basis of full inverse planning model because the model needs all plans of $\mathcal{P}_g$, and this is almost impossible for humans in complex situations. Humans consider several plans to estimate others' goal and they tend to consider plans that they can predict easily. This is the reason that humans tend to think that others take actions according to such unintuitive process and this is plan predictability bias.





\section{Experiment}

\begin{figure}[t]
  \begin{minipage}{\hsize}
  \begin{center}
   \includegraphics[width=0.7\linewidth]{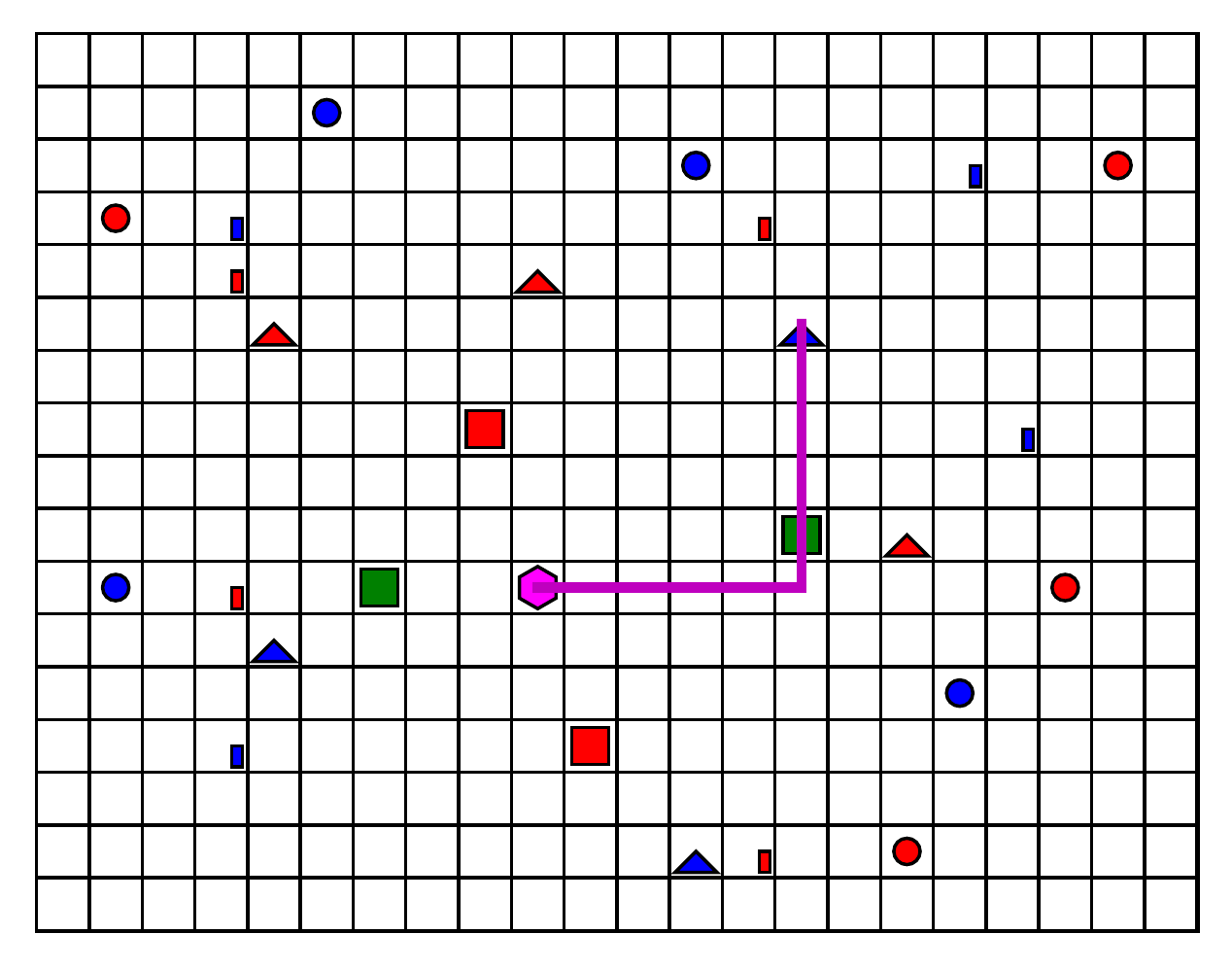}
  \end{center}
 \end{minipage}
 \begin{center}
 \begin{minipage}{0.23\hsize}
  \begin{center}
   \includegraphics[width=\linewidth]{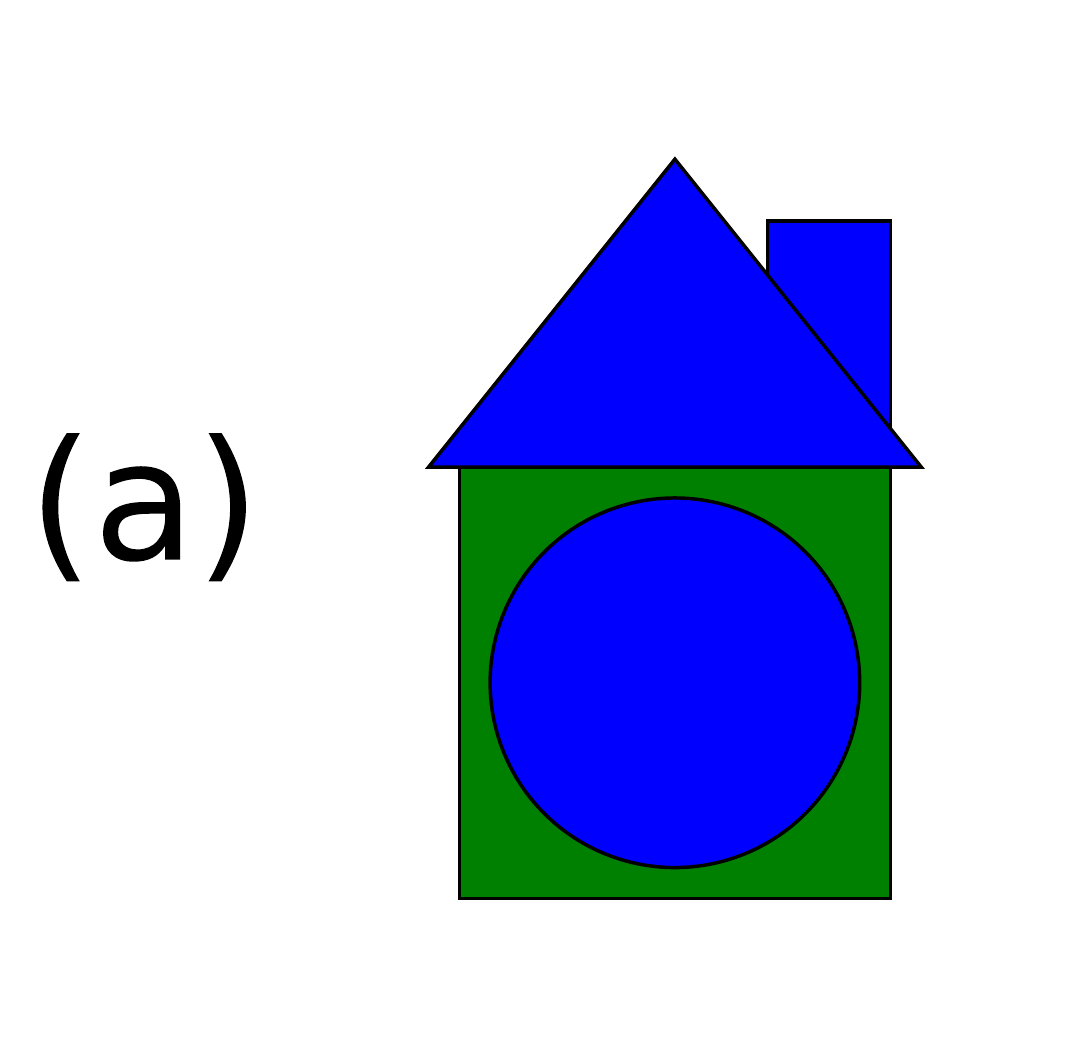}
  \end{center}
 \end{minipage} 
 \begin{minipage}{0.23\hsize}
  \begin{center}
   \includegraphics[width=\linewidth]{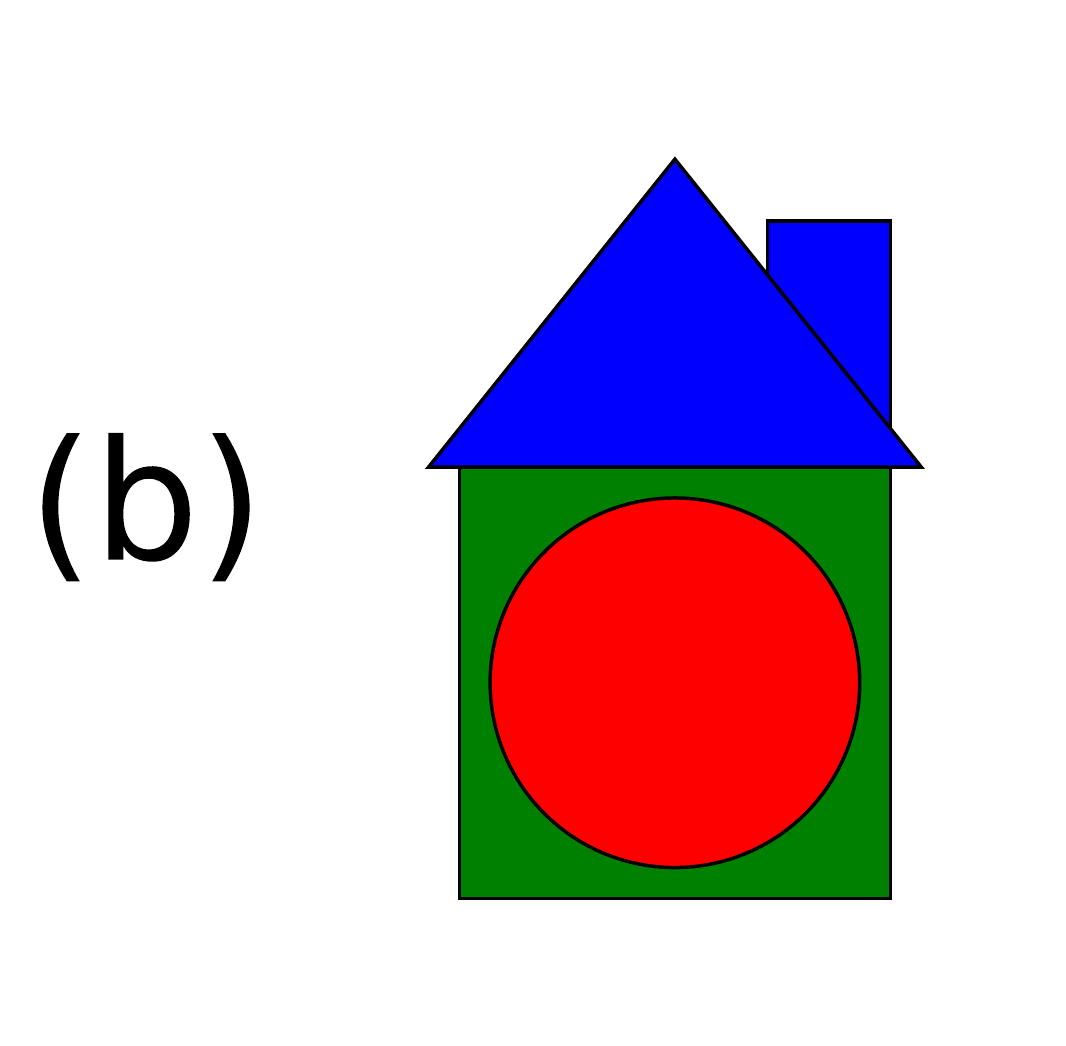}
  \end{center}
 \end{minipage} 
  \begin{minipage}{0.23\hsize}
  \begin{center}
   \includegraphics[width=\linewidth]{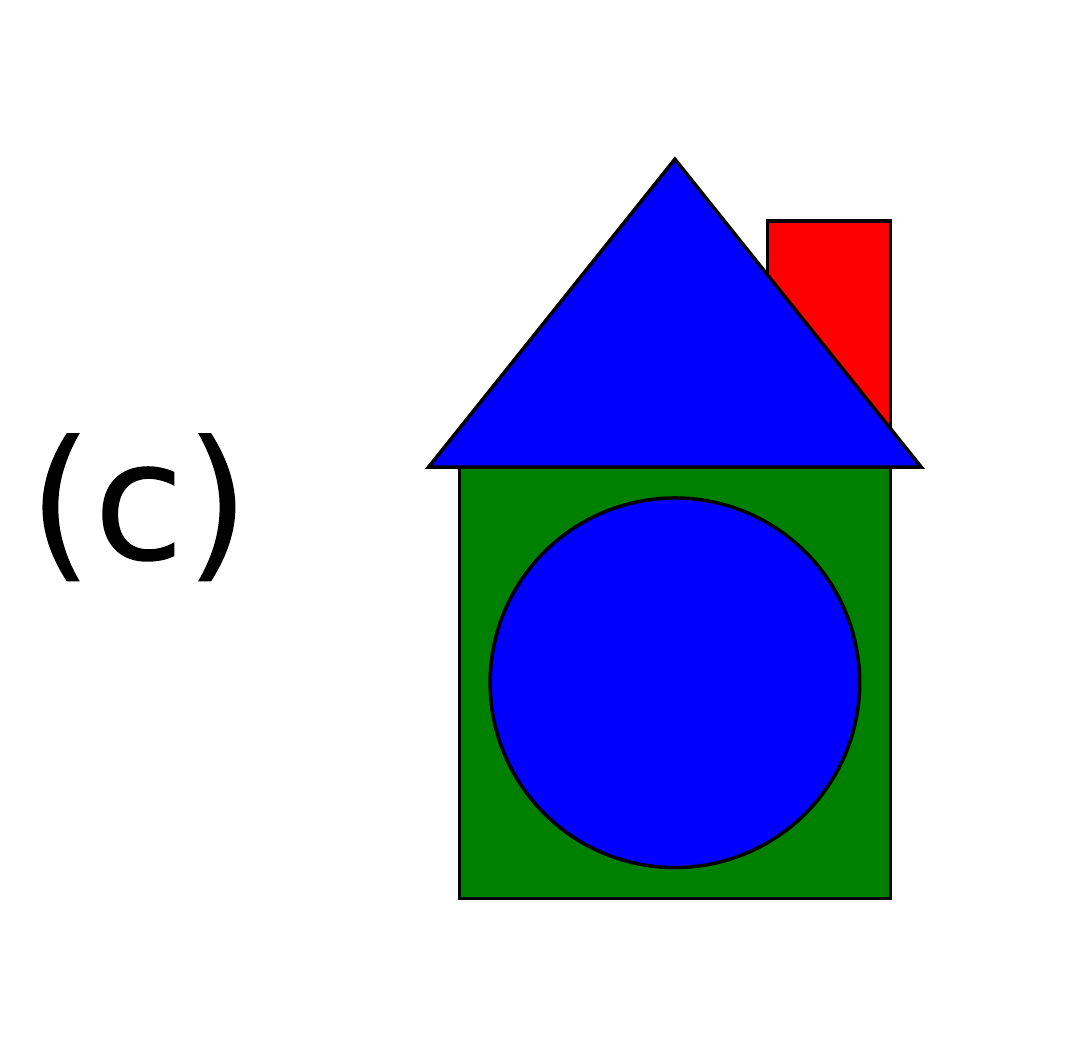}
  \end{center}
 \end{minipage} 
 \begin{minipage}{0.23\hsize}
  \begin{center}
   \includegraphics[width=\linewidth]{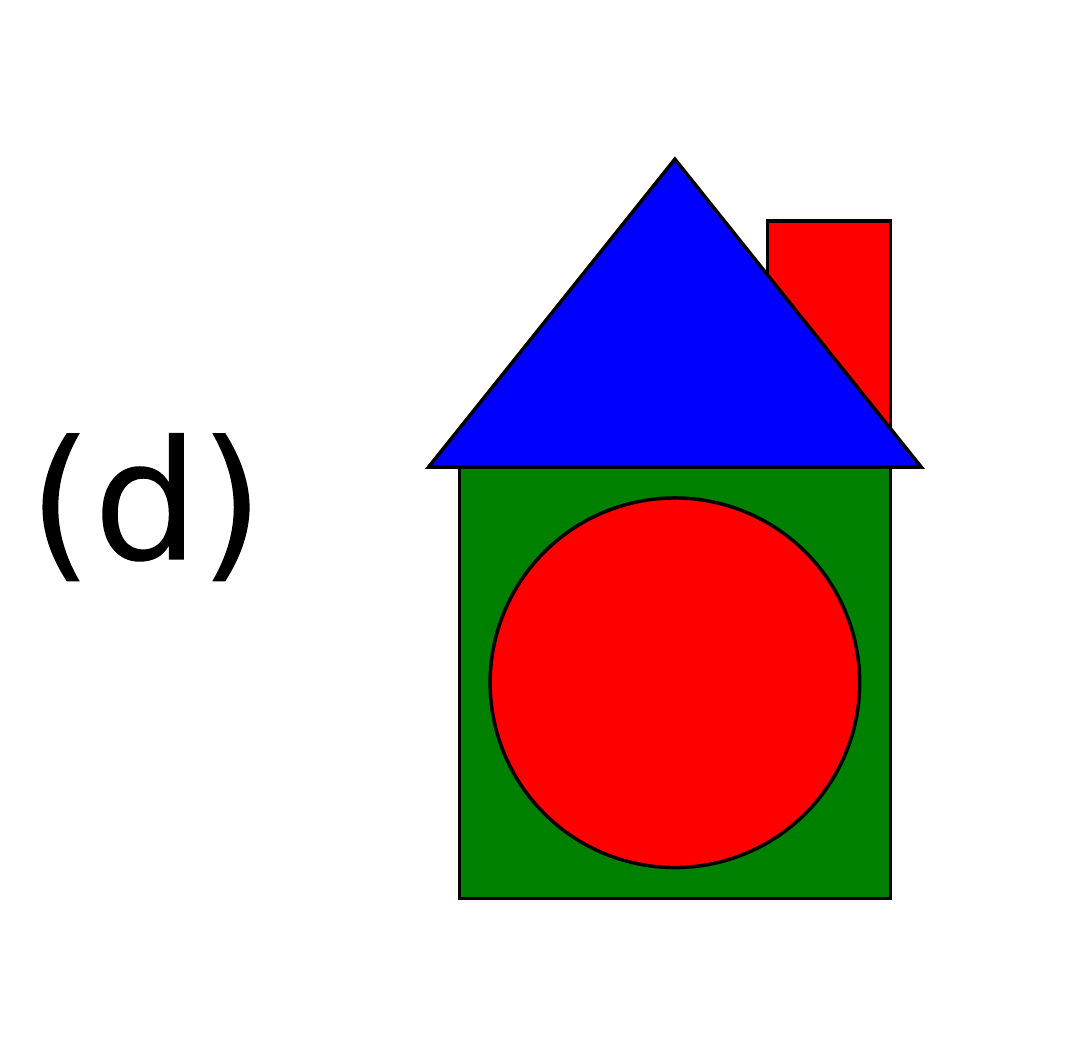}
  \end{center}
 \end{minipage}
 \end{center}
 \caption{Example of ``item creating'' scenarios [for task (4, 2, 2)]. Location of agent is represented as purple hexagon; other red or blue shapes represent parts. Purple line is partial agent path. This means agent got green rectangle and blue triangle. Participants selected and evaluated their inferences toward ``goal product'' that agent wanted to create between lower four candidates.}
 \label{fig:one}
\end{figure}

To compare the full inverse planning model and the plan predictability oriented model with human cognition, we did subject experiments. For these experiments, we considered ``item creating'' scenarios.

In Figure \ref{fig:one}, the upper figure is an example of the scenario. The environment in this scenario is a kind of grid world. There is one agent and several parts of items in either grid, and there is not more than one part in the same grid. Here, the agent is represented by a purple hexagon. There are four types of parts (square, triangle, small rectangle, circle), and there are two to three colors for each type.

The goal of the agent is to create a ``goal product'' that an agent wants to create. The ``goal product'' consists of two to four types of parts with only one type used one time for each product (square, triangle), (square, triangle, small rectangle), (square, triangle, small rectangle, circle). The agent moves to collect the parts that are necessary for its own ``goal product.'' The agent has the priority to collect the items. The agent collects the parts in the order of square, triangle, small rectangle, circle. There are multiple objects of the same color and the same type in the environment; thus, there is a more than one combination of objects for generating one object.


\subsubsection{Participants}
We recruited participants for this study using Yahoo! Cloud Sourcing. Valid participants were 47 adults located in Japan (13 male, 29 female, 5 unknown). The mean age was 39 years old.

\begin{figure}[t]
 \begin{center}
  \includegraphics[width=0.7\linewidth]{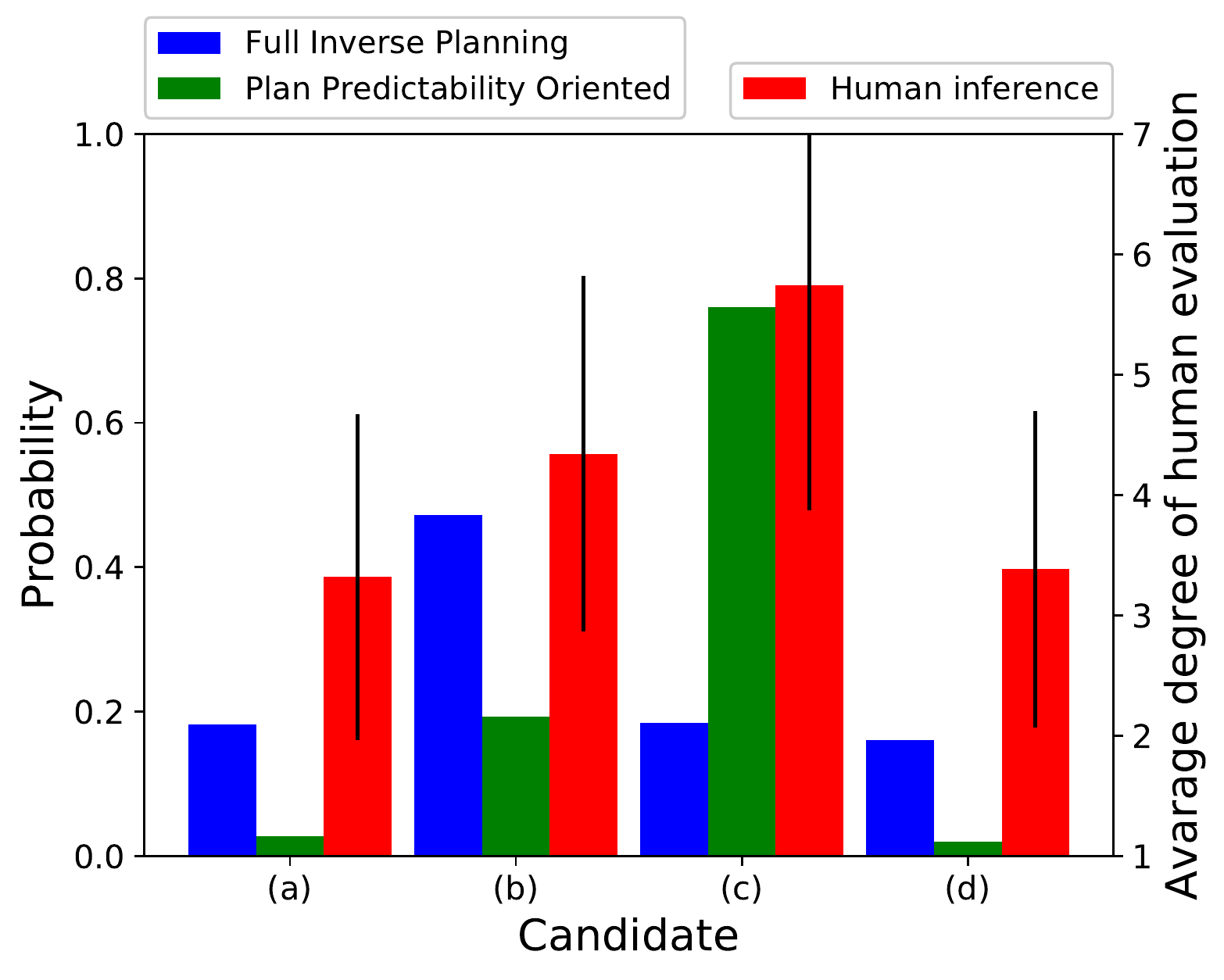}
 \end{center}
 \caption{Inference of human and computational models for task (4, 2, 2)}
 \label{fig:result_4}
\end{figure}

\begin{table}[t]
  \centering
  \begin{tabular}{|c||c|}
    \hline
    Full Inverse Planning Model & 0.765 ($p \ll 0.0001$)\\
    \hline
    Plan Predictability Oriented Model & 0.916 ($p \ll 0.0001$)\\
    \hline
  \end{tabular}
  \caption{Pearson correlation between full Bayesian model and plan prediction oriented model with human inferences}
  \label{table:cor_all}
\end{table}

\subsubsection{Procedure of Experiment}
Experiments were conducted on the Web via a browser application we made.
Subjects were instructed on the rules of agent behavior and then underwent a confirmation test to check their degree of understanding. In this test, participants who were judged to not understand the rules were given the instructions again. The participants who passed the confirmation test entered the actual experiment phase. In this phase, participants saw the environment, part of the agent's movement path for collecting parts, and four target candidates for the agent's ``goal product'' simultaneously. The subjects selected one that they considered most likely to be the agent's ``goal product'' from the candidates. Also, participants scored the degree of likelihood for which they estimated a candidate as being the agentâ€™s ``goal product'' for all the candidates. We adopted a seven-degree score for the evaluation.

Before analyzing the participants' results, we excluded the results of the participants who were invalidated. We defined invalid participants as participants who gave the same evaluation score to all candidates or did not give the highest evaluation score to the selected candidate as the most likely candidate.

\begin{figure}[t]
 \begin{center}
  \includegraphics[width=0.7\linewidth]{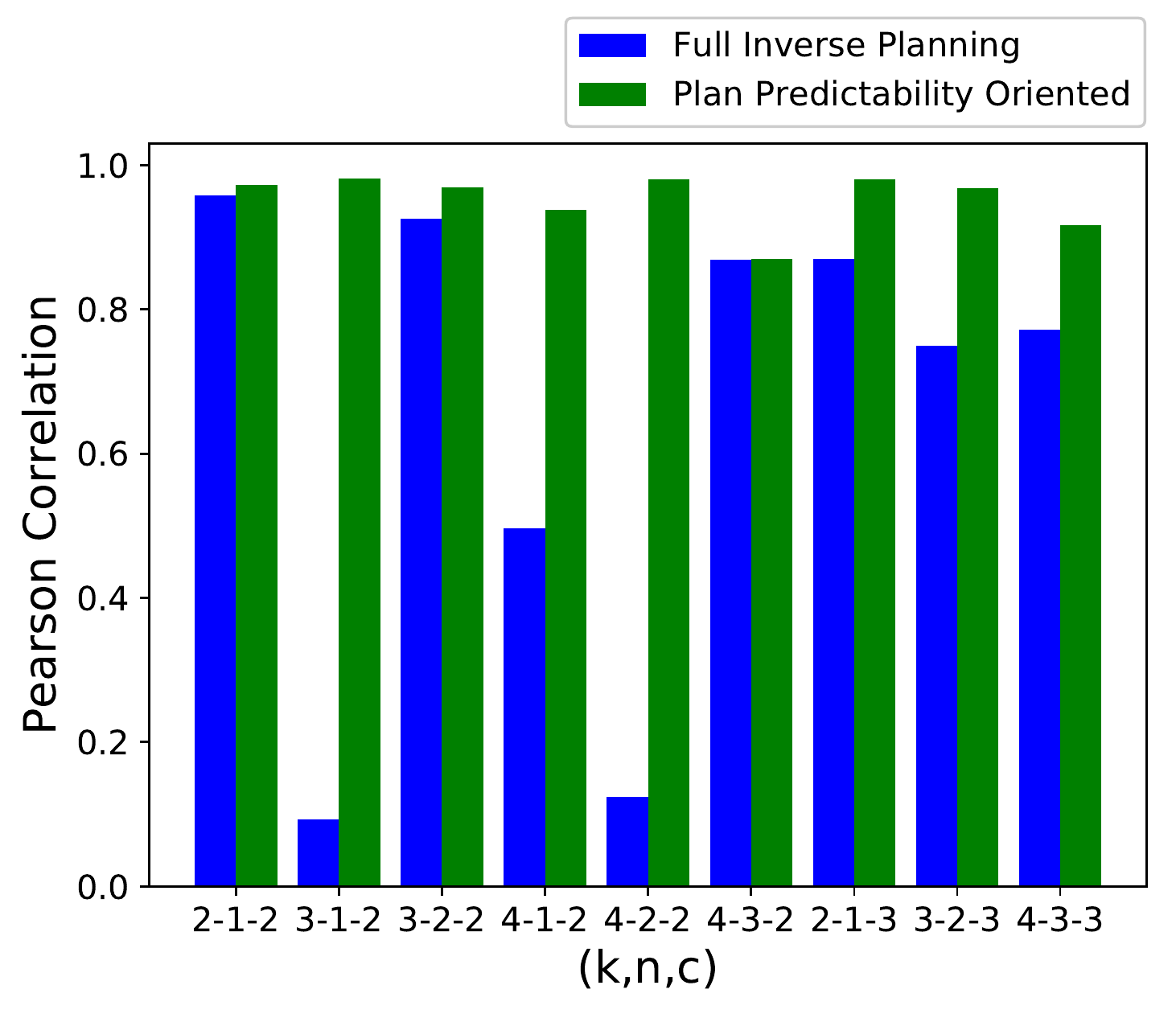}
 \end{center}
 \caption{Pearson correlation between full inverse planning model and plan predictability oriented model with human inferences for each task}
 \label{fig:result_1}
\end{figure}

\begin{table}[t]
  \centering
  \begin{tabular}{|c||c|}
    \hline
    Full Inverse Planning Model & -0.714 ($p = 0.03$)\\
    \hline
    Plan Predictability Oriented Model & 0.116 ($p = 0.76$) \\
    \hline
  \end{tabular}
  \caption{Pearson correlation of ``Pearson correlation of human inference with each models'' with task complexity factor $k-n$}
  \label{table:car_val}
\end{table}

\subsubsection{Stimuli}
We prepared nine stimuli (tasks) with different task complexities. There are three variables that affect task complexity. $k$ is the number of types of parts included in the agent's ``goal product,'' $n$ is the number of types of parts included in the agent's path, and $c$ is the number of colors of parts that have not been collected by the agent yet. We designed nine combinations of complexity for variables $k$, $n$, and $c$. There were (2, 1, 2), (3, 1, 2), (3, 2, 2), (4, 1, 2), (4, 2, 2), (4, 3, 2), (2, 1, 3), (3, 2, 3) and (4, 3, 3). we made a stimulus corresponding to each combination. In task $c = 3$, the only type of part that was not yet collected by the agent had three colors. The example in Figure \ref{fig:one} is task (4, 2, 2), and the purple line in the upper figure corresponds to the movement of the agent.

We designed the placement of parts within the environment and the agentâ€™s path within the task to make the inference of the most likely ``goal product'' different with the full inverse planning model and plan predictability oriented model. We chose four candidates according to four policies: the most likely candidate for the full inverse planning model, the most likely candidate for the plan predictability oriented model, and the candidate that had a low probability for both models.
The lower figures in Figure \ref{fig:one} show example candidates for the agent's ``goal product'' for task (4, 2, 2).

\subsubsection{Model}
In the ``item creating'' scenario, $\mathcal{G}$ corresponds to a set of ``goal products'', and $\mathcal{A}$ corresponds to a set of individual parts.
$\mathcal{P}_g$ is a set of all available combinations of parts to build a ``goal product'' $g$. Since there is a more than one combination of objects for generating one object, $\forall g, |\mathcal{P}_g| > 1$.
Here, we defined $Q_g(p)$ as $-cost(p)$. $cost(p)$ is the shortest path length of $p$. 
$Q_p(\mathbf{a})$ is $-cost(p - \mathbf{a})$. $p - \mathbf{a}$ means the remaining plan of $p$ after $\mathbf{a}$.
We set rational parameters as $\beta_1 = 0.3$, $\beta_2 = 0.3$, $\beta_3 = 0.5$. 

\subsubsection{Result}

We evaluated the full inverse planning model and plan predictability oriented model with a comparative experiment by comparing participant scores. 
For comparison, we made a human score vector and model probability vector. The human score vector was a vector that consisted of the participants' scores serialized over all results (thus, the length of the vector was 36) without any normalization. We used the average vector of human score vectors for all valid participants. To make the model probability vector, we extracted probabilities of candidates for each task and serialized them (the length of the vector was also 36). Table \ref{table:cor_all} is a Pearson correlation of the averaged human score vector between model probability vectors for both models. The results show that the plan predictability oriented model had a much better correlation with human inference. Figure \ref{fig:result_4} is the specific result for task (4, 2, 2). Blue and green are the probability calculated by each computational model, and the red bar is the average of the participants' scores. This figure also shows a good correlation of human inference with the plan predictability oriented model. We also executed a significance test. First, we calculated the Pearson correlation between human score vectors and model probability vectors for all valid participants. Thus, we obtained two sets of Pearson correlations for the two models. Then, we executed t-tests on the sets. The p-value was 0.03 ($<$ 0.05). Thus, we confirmed that there was a significant difference in correlation between the two models. 

\paragraph{Relation to task complexity}
Next, we calculated the Pearson correlation of human inference with both models for each task. We made an averaged human score vector and model probability vectors for each task and calculated the Pearson correlation by using these vectors. Figure \ref{fig:result_1} is the result. First, the results show that the plan predictability oriented model had a much better correlation with human inference for all of the tasks. The full inverse planning model had a low correlation with human inference for tasks for (3-1-2), (4-1-2), and (4-2-2) in particular. The common factor in these tasks was that the remaining number of types of parts, which is represented as $k-n$, was more than one.

Table. \ref{table:car_val} is the Pearson correlation of ``Pearson correlation of human inference with each model'' with $k-n$. In other words, it is the correlation between the values of Figure \ref{fig:result_1} and $k-n$. The full inverse planning model had a strongly negative correlation with $k-n$. 
$k-n$ was strongly related to the future available paths of the agent. This means that the full inverse planning model was not effective for tasks that had many future available paths. This matches with the intuition that humans may think bounded-rationally, not full-rationally, in complex situations. The plan predictability oriented model did not have such negative correlation with $k-n$. This means that this model was not affected by task complexity.

\paragraph{Confirmation of individual difference}
\begin{figure}
 \begin{center}
  \includegraphics[width=0.7\linewidth]{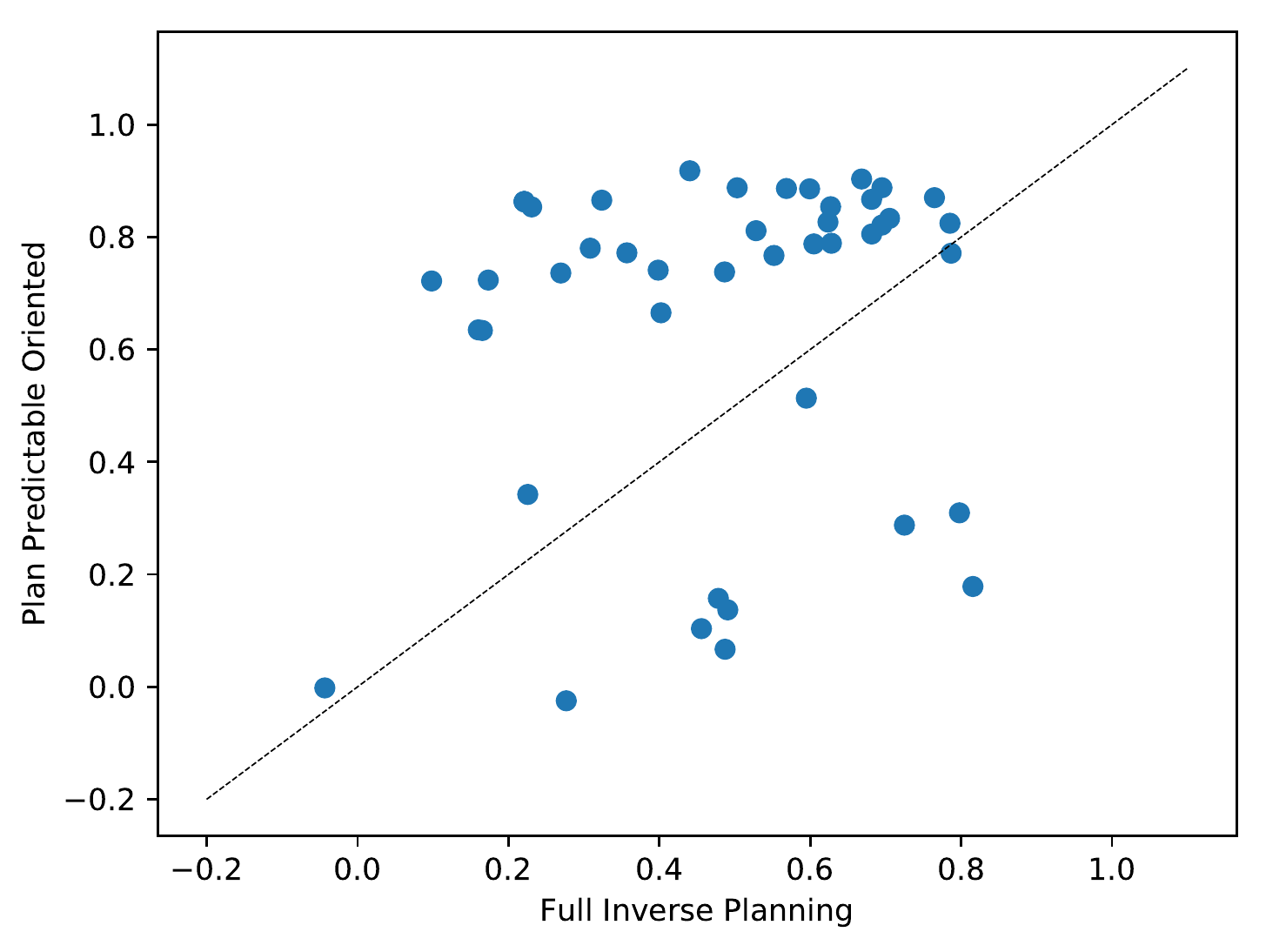}
 \end{center}
 \caption{Scatter plot of Pearson correlation between full inverse planning model and plan predictability oriented model with human inference for each participant}
 \label{fig:result_2}
\end{figure}

We calculated the Pearson correlation of human inference with both models for each participant. Figure \ref{fig:result_2} is a scatter plot of the results. The dotted line on the plot shows the boundary where the correlation between both models was equal. Most of the participants are on the upper left of the plot. This means that most of the participants' inference has a good correlation with the plan predictability oriented model. However, some participants on the lower right of the plot means that the plan predictability oriented model cannot model for these participants. The participants could rationally recognize other peopleâ€™s intentions completely in this experiment, so they had a good correlation with the full inverse planning model. These results suggest that there are individual differences in peoplesâ€™ bounded rationality.

\begin{figure}
 \begin{center}
  \includegraphics[width=0.7\linewidth]{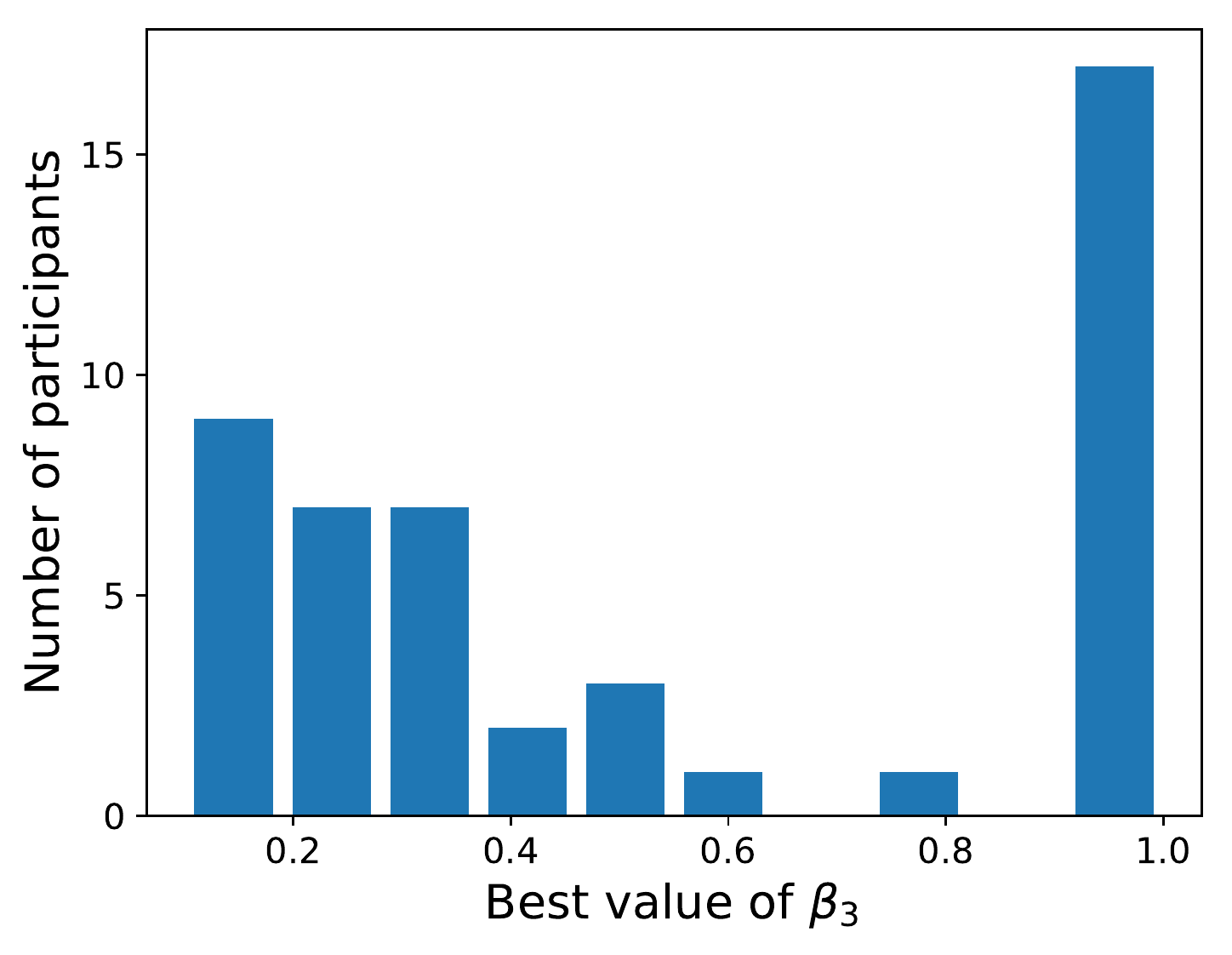}
 \end{center}
 \caption{Histograms for the number of participants for each best predictability bias}
 \label{fig:result_3}
\end{figure}

We made multiple plan predictability oriented models that had different $\beta_3$. $\beta_3$ is the parameter for plan predictability bias. The range of $\beta_3$ was from 0.0 to 1.0 in increments of 0.1. Figure \ref{fig:result_3} is a histogram of the number of participants who had the best correlation with the plan predictability oriented model with $\beta_3$. It shows that many of the participants had a strong bias, but some had a small bias or no bias. Table \ref{table:inidiv} is the average Pearson correlation for the result of each participants with the models. Here, (same) means that we used the same value for $\beta_3$, and (individual) means we used the best value for $\beta_3$ for individual participants. The individual setting had a higher correlation with humans than the same correlation. This suggests that adaptating the bias can improve our model.

\begin{table}
  
  \centering
  \begin{tabular}{|c||c|}
    \hline
    Full Inverse Planning Model & 0.513 \\
    \hline
    Plan Predictability Oriented Model (same)& 0.638 \\
    \hline
    Plan Predictability Oriented Model (individual)& 0.738 \\
    \hline
  \end{tabular}
  \caption{Average Pearson correlation of human inference between models for each participant}
  \label{table:inidiv}
\end{table}

\section{Discussion}

Essentially, full inverse planning model differs from human cognition in situations in which there is a difference in the rationality of the actor and the observer. Since forward planning for taking action towards a particular goal is generally easier than inverse planning to infer cause from actions observed, these situations might always happen. Therefore, we think that the model is useful for many situations. In addition, our computational model is based on the inverse planning model and simple plan prediction, so it has the potential of being adapted to various situations. However, we showed the actual effectiveness only under the ``item creating'' scenario. This scenario is one example environment that has a grid geometric rationality and sequential planning. Explicit sequential planning is the most basic planning, and a lot of human planning is based on sequential planning. The grid geometric rationality and sequential planning are often used in the theory of mind \cite{Baker2017}. Therefore, we think that our model can be used in broad situations.


Understanding how humans infer others' intentions is useful for considering good actions when corroborating with others. In the cognitive science area, there is research on how humans behave when they want to communicate their goal or purpose  \cite{Shafto2014}. In the artificial intelligence and robotics areas, research on collaborative planning is more popular and important. For example,``legibility'' is proposed \cite{Dragan2014}. This is a measure of human expectation toward robots' intentions or goals as based on the behavior of robots. There are also works on using ``legibility'' for planning \cite{Fisac2017}.

The expansion of our model for larger and more complex tasks is a very interesting direction for our future work. Introducing hierarchical planning is a promising approach. Our model can be considered as one type of hierarchical modeling in which the inference of plans is regarded as an intermediate layer. The hierarchical predictive coding framework \cite{Blokpoel2012} is one example of hierarchical modeling for human cognition. This model has multiple inference layers with different abstraction levels, and execute step-by-step inference by using MAP estimation.
Similarly, our model can be expanded with multiple planning layers. Determining whether such a model is better for modeling human cognition would be an interesting next research step. 



Deeper analysis of individual rationality is also interesting. We just demonstrated that human rationality differs from person to person. However, there might be some factors that decide the degree of bias. Seeking such factors and improving our model by implementing them would be a valuable study.

\section{Conclusion}

In this paper, we proposed a novel computational model called the ``plan predictability oriented model'' to infer the goals of others through their behavior. This model implements bounded rationality for complex tasks that have many options for one purpose. We confirmed that our model has a better correlation with human inference than the full inverse planning model via a subject experiment using the “item creating” scenario. We also confirmed that the full inverse planning model becomes progressively worse with the increasing complexity of tasks, while our model remains unaffected by changes in complexity. This suggests that our model has robustness for complex tasks. We also confirmed the existence of individual differences in “bounded rationality” and suggested that we could improve our model by introducing individualized bounded rationality.

Although there are many limitations and much room for improvement, the model is valuable as one example of the theory of mind with bounded rationality. We are confident that this result can contribute to research on human cognition and the development of engineering applications under cognitive science.

\section{Acknowledgements}
This study was partially supported by JSPS KAKENHI ''Cognitive Interaction Design'' (No.JP26118005).

\bibliographystyle{apacite}

\setlength{\bibleftmargin}{.125in}
\setlength{\bibindent}{-\bibleftmargin}

\bibliography{main}

\end{document}